\documentclass{edm_article}

\usepackage{graphicx}
\usepackage{xcolor}
\usepackage{amsmath}
\usepackage{multirow}
\usepackage{multicol}
\usepackage{booktabs}
\usepackage{enumitem}
\usepackage{hyperref}

\begin{document}

\title{Can Large Language Models Replicate ITS Feedback on Open-Ended Math Questions?}

\numberofauthors{5}
\author{
Hunter McNichols$^1$, Jaewook Lee$^1$, Stephen Fancsali$^2$, Steve Ritter$^2$, Andrew Lan$^1$\\
       \affaddr{University of Massachusetts Amherst$^1$, Carnegie Learning$^2$}\\
       \email{wmcnichols@umass.edu}
}

\maketitle

\begin{abstract}
Intelligent Tutoring Systems (ITSs) often contain an automated feedback component, which provides a predefined feedback message to students when they detect a predefined error. To such a feedback component, we often resort to template-based approaches. These approaches require significant effort from human experts to detect a limited number of possible student errors and provide corresponding feedback. This limitation is exemplified in open-ended math questions, where there can be a large number of different incorrect errors. In our work, we examine the capabilities of large language models (LLMs) to generate feedback for open-ended math questions, similar to that of an established ITS that uses a template-based approach. We fine-tune both open-source and proprietary LLMs on real student responses and corresponding ITS-provided feedback. We measure the quality of the generated feedback using text similarity metrics. We find that open-source and proprietary models both show promise in replicating the feedback they see during training, but do not generalize well to previously unseen student errors. These results suggest that despite being able to learn the formatting of feedback, LLMs are not able to fully understand mathematical errors made by students.\footnote{Source code is available at \url{https://github.com/umass-ml4ed/its_feedback_edm}} 
\end{abstract}


\begin{figure}[t!]
    \Description{An open-ended math question with an ITS's built-in, template-based feedback for an incorrect student response.}
    \centering
    \includegraphics[width=0.99\linewidth]{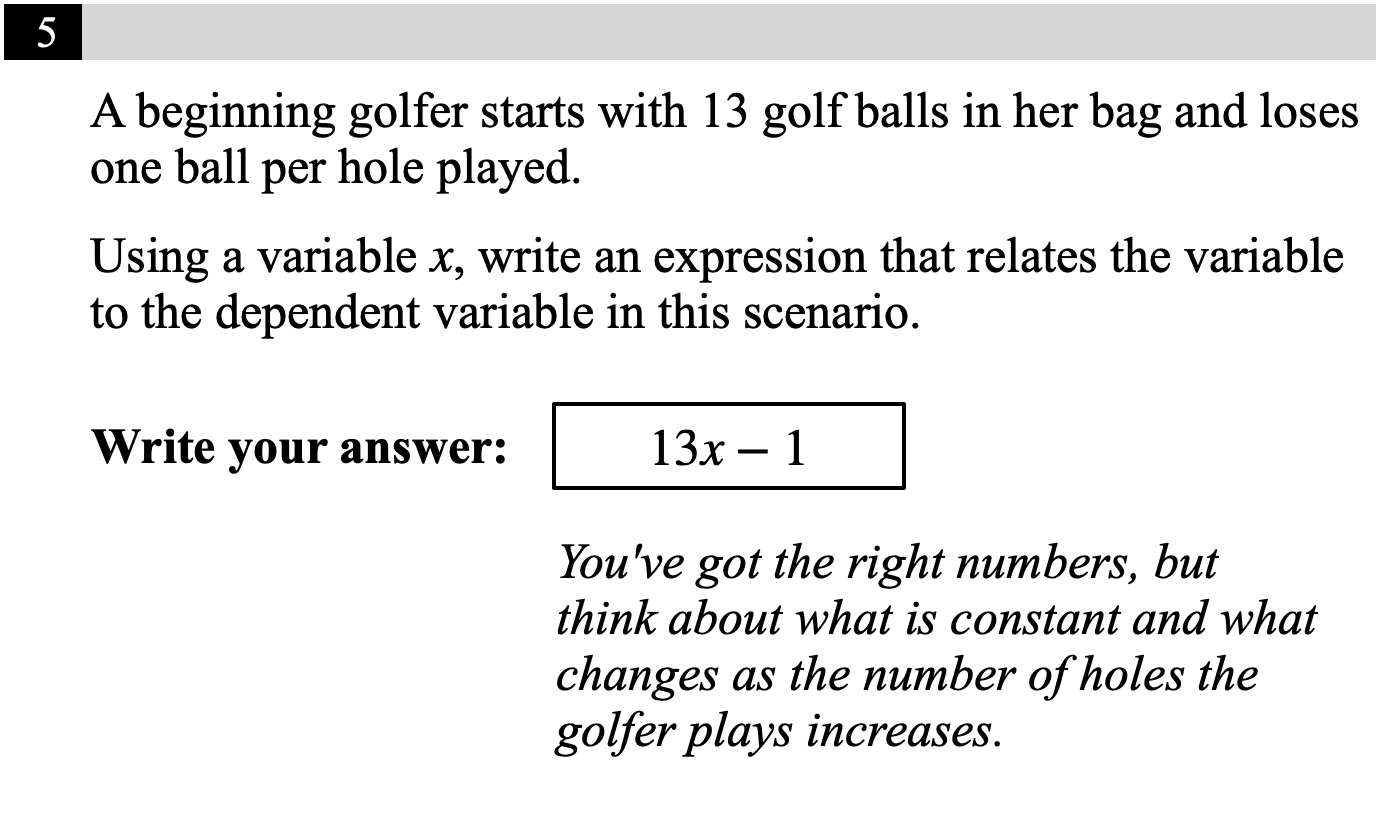}
    \caption{An open-ended math question with an ITS's built-in, template-based feedback for an incorrect student response.}
    \label{fig:question-example}
\end{figure}
\section{Introduction}

High-quality math education is increasingly important in today's world since it is highly relevant in many science, technology, engineering, and mathematics (STEM) subjects. One effective way to scale high-quality math instruction to a large number of \emph{students} is through intelligent tutoring systems (ITSs) and online learning platforms, which provide learning opportunities to students both in-class and on demand. One of the major advantages of ITSs is that they can provide immediate feedback to students when they respond to assignment questions incorrectly~\cite{stamper2008hint}, which has been shown to improve learning outcomes \cite{feedback-effect-2, feedback-effect-1}.

Historically, the majority of feedback components in ITSs rely on a rule-based approach \cite{feedback-gen-neil, lan2015mathematical}. In this approach, math education experts partner with ITS developers to 1) anticipate common student errors, either from past experience or actual student data, 2) connect them to resulting incorrect responses to questions, and 3) develop corresponding feedback, often in the form of textual feedback messages, which are automatically deployed to students whose response corresponds to one of the anticipated errors. Doing so is especially challenging for open-ended math problems compared to true/false or multiple-choice ones, since there can be a large number of possible ways for students to make errors, which may all lead to different incorrect responses \cite{gurung2023common}. 

While such feedback approaches are reliable and commonly used in large-scale ITSs, they are limited by their hand-coded nature. First, these messages can only be shown if the student makes an anticipated error, and cannot account for errors not predicted by the content developers before deployment. Second, when developing feedback for a new question that does not correspond to any of the existing templates, content developers have to manually craft these feedback messages, which is labor-intensive and limits the scalability of ITSs. Therefore, in order to scale an ITS system to a larger amount of \emph{content}, questions in particular, we need to explore methods for automated feedback generation. Recent, state-of-the-art large language Models (LLMs) show impressive capabilities in generating fluent text under textual instructions and even mathematical reasoning abilities, which raises the question: can we use LLMs to automatically generate feedback to incorrect student responses, or to take a step back and be more restrictive, at least \emph{replicate existing ITS feedback mechanisms?}

\subsection{Contributions}
In this work, we explore the capabilities of LLMs to replicate the built-in feedback mechanisms of ITSs, by automatically generating feedback messages to students' incorrect responses to open-ended math questions. \textbf{We limit the scope of our work to rule-based feedback mechanisms involving hand-crafted templates}; replicating human-authored, free-form feedback \cite{haim2022toward,mcnichols2023exploring} in ITSs is left for future work, due to the significant variability in styles and content in those feedback messages \cite{baral2023investigating}, even for the same question-response pair. First, we formulate the automated feedback generation problem and adopt several well-studied methods for this task. Second, we perform extensive experimentation on a training dataset that consists of real student responses and feedback messages from a large-scale math ITS system. We investigate both open-source LLMs and proprietary LLMs across multiple experimental settings. Our results show that LLMs can replicate highly structured feedback given appropriate training data, but cannot generalize to previously unseen errors. These results suggest that LLMs are more capable of capturing the structure of text than understanding how math errors occur among students.

\section{Related Works}

The last several years have seen increasing interest in how LLMs can be used to generate feedback, through prompting-based approaches \cite{socratic, distractors, feedback-gen-decimal, feedback-gen-chatgpt} or with fine-tuning \cite{insta-reviewer, koutcheme2023training}. However, none of these works have explored using (relatively large) open-source LLMs, such as Mistral-7B \cite{jiang2023mistral},  derived from Meta's recently released open-source Llama 2 model \cite{touvron2023llama} which we study extensively in this paper. 

The approach of using an LLM to generate feedback is appealing due to the convenience and ease of prompting them in comparison to more manual, rule based-approaches, but there remains a question about the quality of this feedback. In the math domain, existing work has shown that there still appears to be a considerable gap in quality between teacher-authored feedback and LLM-authored feedback~\cite{prihar2023comparing}. 

\section{Approach}

We now define the task of feedback generation and our approach to using LLMs for this task.

\subsection{Feedback Generation Task}
Consider the example open-ended math question on linear equations and a student response shown in Figure~\ref{fig:question-example}. In this example, the student correctly defines a variable and linear equation but submits an incorrect response by switching the slope with the intercept. The feedback message provided by an ITS provides a hint to the student and reminds them of the key intuition behind linear equations. At the same time, the message avoids directly revealing the correct response and intends to guide the student towards working it out themselves. Our goal is to automatically generate such a feedback message using an LLM. Formally, we define feedback generation as the task of generating such a feedback $m$, given a question body $q$, the correct response $c$, and an erroneous student response $a$.

We treat each aforementioned component in the problem as a series of tokens, e.g., the feedback message is defined as $m = \{x_1, x_2 \dots x_M\}$, where $M$ is the total number of tokens in the message. Large-scale decoder-only LLMs are trained to predict the next token probability distribution given an existing sequence of tokens. Therefore, we frame the feedback generation task in terms of sampling an output $m$ from an LLM token-by-token, given the input sequence $q$, $c$, and $a$, and possibly some textual statements that provide additional instructions on how to generate feedback. We refer to this input as the prompt. This process is summarized as
\begin{equation*}
    (q, c, a, x_1, \dots x_{i-1}) \xrightarrow{LLM} x_i,
\end{equation*}
where $i$ indexes tokens in the feedback message.

\begin{figure}
    \centering
    \Description{Example prompt provided to an LLM for feedback generation. The shaded information is an in-context example to guide the model to produce output with similar structure.}
    \includegraphics[width=0.99\linewidth]{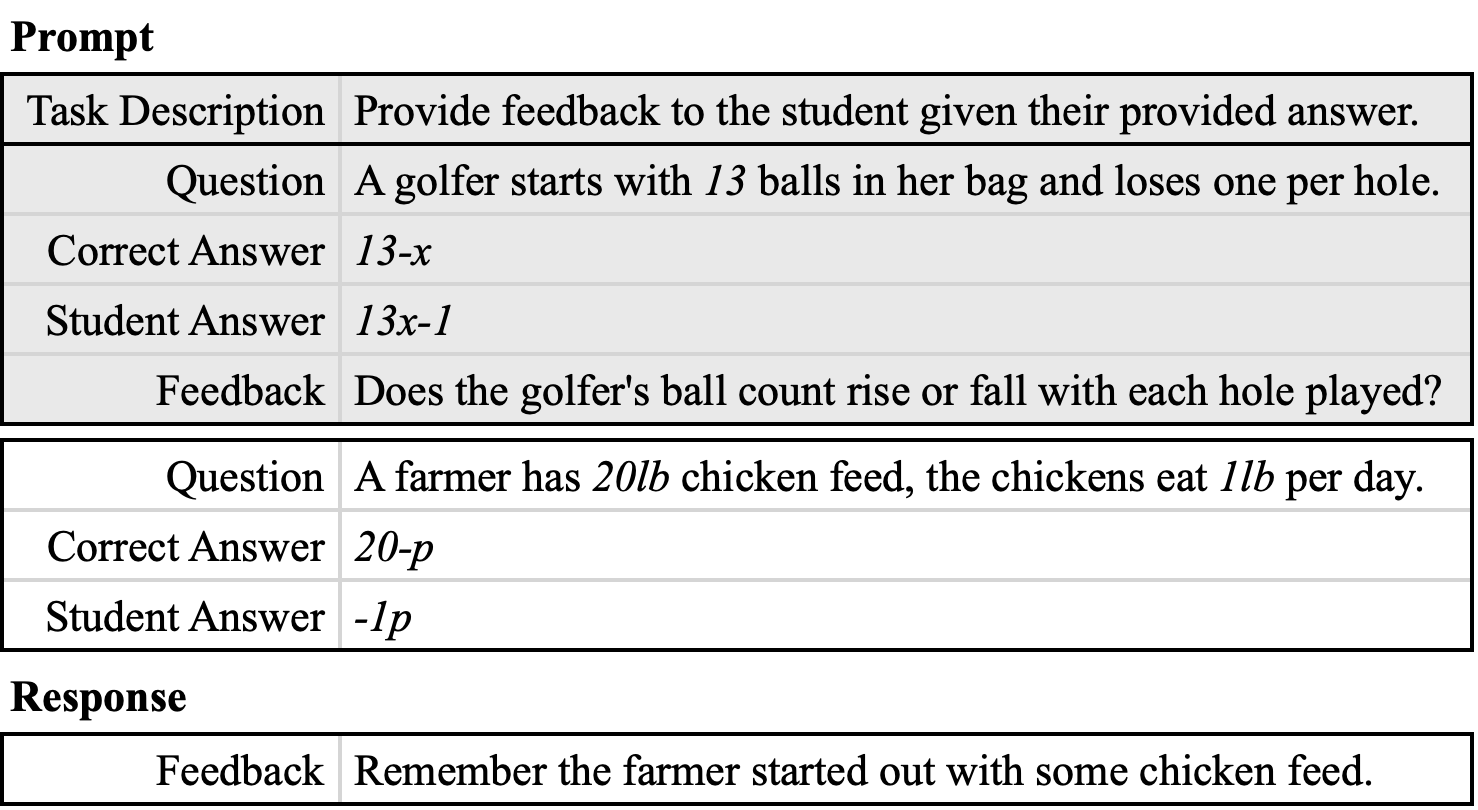}
    \caption{Example prompt provided to an LLM for feedback generation. The shaded information is an in-context example to guide the model to produce output with similar structure.}
    \label{fig:prompt-example}
\end{figure}

\sloppy
\subsection{Fine-Tuning and In-Context Learning}
To test whether LLMs can replicate the built-in feedback mechanisms in ITSs, we can use a training set that consists of $(q, c, a, m)$ tuples to align the LLM's output with built-in feedback messages. A simple way to do so is fine-tuning, which directly updates the LLM's parameters (in practice, usually a subset of them) using this training data. To do this, we minimize the negative log likelihood that the given ``ground truth'' feedback message $m$ is generated given the input $q,c,a$, by summing over all tokens in the feedback message, i.e., 
\begin{equation*}
    \frac{1}{N}\sum_{i=1}^N \frac{1}{M_i} \sum_{j=1}^{M_i} -\log p(x^i_j | q^i, c^i, a^i, x^i_1, \ldots, x^i_{j-1}),
\end{equation*}
where $N$ is the number of training tuples/data points, $q^i, c^i, a^i$ denote the question body, correct response, and incorrect student response in the $i^\text{th}$ training sample, and $x^i_j$ is the $j^\text{th}$ token in the corresponding feedback message (with a total of $M_i$ such tokens). This way, we modify the LLM's parameters so that its output token distribution is closer aligned with existing feedback messages. 

\sloppy
Fine-tuning involves modifying the LLM's parameters, which can be computationally expensive and can be hard to do for models that have billions of parameters. Alternatively, another approach to replicate built-in feedback is called In-Context Learning (ICL)~\cite{incontext2021gpt3}. ICL is performed by including examples of the desired prompt-message token sequence directly in the LLM's input prompt, as shown in Figure~\ref{fig:prompt-example}. The inclusion of such ``in-context'' examples, which provide LLM with a highly specific demonstration of the task structure, also modify its token distribution to align it with existing feedback messages. This approach has been shown to be highly effective for the largest LLMs and achieves state-of-the-art performance on many natural language generation tasks~\cite{brown2020language}. We can also combine ICL with fine-tuning by providing the fine-tuned LLM with in-context examples, further improving generation performance.

\section{Experiments}

We now detail the experiments we conducted to test whether LLMs can replicate ITS feedback. We first detail the dataset used, the experimental settings, the metric we use for evaluation, and finally all results and corresponding takeaways. 

\subsection{Dataset}
We utilize a dataset of common erroneous student responses to middle school math questions and corresponding feedback messages in a large-scale ITS. In total, the dataset contains 26,845 unique responses across 100 questions, which are all incorrect responses to open-ended math questions. All questions are word problems that ask students to define an equation for a linear relationship, similar to the one in Figure~\ref{fig:question-example}. Moreover, the dataset also includes the question statement and 
the feedback messages built into the ITS for each erroneous response. The feedback messages are template-based and correspond to student errors that math educators can anticipate, which are deployed when a student enters an anticipated erroneous response for just-in-time feedback. 
Furthermore, the dataset contains a label for each common erroneous response that defines the error the student made. In total, there are 11 unique error labels, yielding 1100 unique feedback messages. For example, the error in Figure~\ref{fig:question-example} has the label \textit{swapped slope and intercept}.

\subsection{Experimental Settings}
We now detail the settings we used for both fine-tuning and in-context learning in our feedback generation experiments.

\sloppy
\subsubsection{Fine-Tuning Setup}

We fine-tune open-source LLMs by training a Low Rank Adaptation (LoRA) instead of directly modifying the weights~\cite{hu2021lora}. This approach trains a low-rank adaptor for each weight matrix in the LLM architecture instead of directly adjusting the weights. This approach allows us to train larger models on more accessible hardware configurations. We use the AdamW optimizer~\cite{loshchilov2019decoupled} with a learning rate of $10^{-5}$ and LoRA hyperparameters of $r=16$ and $\alpha=16$. For open-source LLMs, we utilize the \textit{Mistral-7B-v0.1} model sourced from the HuggingFace library. We select Mistral-7B due to its reported superior performance over other LLMs on math reasoning tasks~\cite{jiang2023mistral}. For proprietary models, we select ChatGPT~\cite{chatgpt}, specifically \textit{gpt-3.5-turbo-1106} which is the most powerful model available from OpenAI with fine-tuning enabled. For fine-tuning, via OpenAI's API, we use the default training hyperparameters.

We utilize only a small percentage of the available dataset in order to fine-tune our models. Specifically, we constrain our training sets to be 100 feedback examples, which is the recommended best practice provided by OpenAI in their fine-tuning documentation. In our experiments, we found that this suggestion applies to proprietary LLMs as well: when directly fine-tuning Mistral-7B with the entire 1100 examples in the dataset, the model quickly over-fits on the training data within 20\% of the first epoch, showing no signs of generalizability, while using a small, subsampled training set works well. 



\subsection{Evaluation Metrics}

To quantitatively assess how similar the LLM-generated feedback is to the ground-truth, template-based feedback in the ITS, we utilize two reference-based similiarity metrics. First, we compute the bilingual evaluation understudy (BLEU) score~\cite{papineni2002bleu}, which measures the precision of n-gram overlap between the LLM-generated feedback and the ITS's feedback.  Second, we compute Recall-Oriented Understudy for Gisting Evaluation (ROUGE) score~\cite{lin-2004-rouge}, which measures the n-gram recall between the feedbacks. Specifically, we compute BLEU on 4-gram sequences and ROUGE-L, which measures the recall on the longest common sub-sequence. We report the average BLEU/ROUGE score for all feedback messages in the test set; since we have different experimental settings with different train-test splits, we detail their construction in Section~\ref{sec:results}.

\subsection{Methods Compared}
We compare the following variants of fine-tuning and ICL methods for feedback generation, using both the base version and fine-tuned Mistral-7B and GPT-3.5 models:
\begin{itemize}[noitemsep,topsep=0.0pt]
  \item \textbf{Zero}: We use zero-shot prompting and directly instructs the LLM to generate feedback.
  \item \textbf{ICL} (in-context learning): We provide the LLM a single, fixed in-context example and then ask it to generate feedback according to this style.
  \item \textbf{ICL-SE} (ICL with same error): We provide an in-context example that has the same error label as the student response for which we generate feedback.
\end{itemize}

\subsection{Experimental Results and Discussion}
\label{sec:results}
In this section, we detail the results of three experiments where we perform a train-test split across three different dimension of the dataset: response, question, and error.

\begin{table*}[ht!]
  \centering
  \scalebox{0.95}{
  \begin{tabular}{cclccccc}
    \toprule
    \multirow{2}{*}{Split}    & \multirow{2}{*}{Model} & \multirow{2}{*}{Variant} & \multicolumn{2}{c}{Base}                         & \multicolumn{2}{c}{Fine-Tune}                    \\ \cmidrule(lr){4-5} \cmidrule(lr){6-7}
                              & \multicolumn{2}{l}{}                         &  BLEU ($\pm$ STD) & ROUGE ($\pm$ STD) & BLEU ($\pm$ STD) & ROUGE ($\pm$ STD) \\ 
    \midrule
    \multirow{6}{*}{Response} & \multirow{3}{*}{Mistral-7B}       & Zero        & $0.001 \pm 0.0004$      & $0.076 \pm 0.0030$     & $0.164 \pm 0.0158$      & $0.325 \pm 0.0223$     \\
                              &                                & ICL         & $0.077 \pm 0.0062$      & $0.254 \pm 0.0086$     & $0.240 \pm 0.0087$      & $0.448 \pm 0.0084$     \\
                              &                                & ICL-SE      & $0.159 \pm 0.0193$      & $0.307 \pm 0.0244$     & $0.342 \pm 0.0496$      & $0.464 \pm 0.0470$     \\ \cmidrule(lr){3-7}
                              & \multirow{3}{*}{GPT-3.5}        & Zero        & $0.013 \pm 0.0019$      & $0.173 \pm 0.0041$     & $0.285 \pm 0.0298$      & $0.485 \pm 0.0271$     \\
                              &                                & ICL         & $0.044 \pm 0.0037$      & $0.253 \pm 0.0053$     & $0.215 \pm 0.0323$      & $0.449 \pm 0.0265$     \\
                              &                                & ICL-SE      & $0.081 \pm 0.0814$      & $0.311 \pm 0.0074$     & $0.509 \pm 0.0278$      & $0.700 \pm 0.0196$     \\ 
    \midrule
    \multirow{4}{*}{Question} & \multirow{3}{*}{Mistral-7B}       & Zero        & $0.002 \pm 0.0007$      & $0.080 \pm 0.0037$     & $0.128 \pm 0.0274$      & $0.296 \pm 0.0376$     \\
                              &                                & ICL         & $0.080 \pm 0.0063$      & $0.259 \pm 0.0170$     & $0.111 \pm 0.0281$      & $0.349 \pm 0.0465$     \\
                              &                                & ICL-SE      & $0.169 \pm 0.0252$      & $0.333 \pm 0.0295$     & $0.337 \pm 0.0472$      & $0.497 \pm 0.0449$     \\  \cmidrule(lr){3-7}
                              & GPT-3.5                         & ICL-SE      & $0.086 \pm 0.0121$      & $0.318 \pm 0.0107$     & $0.502 \pm 0.0367$      & $0.695 \pm 0.0231$     \\ 
    \midrule
    \multirow{3}{*}{Error}    & \multirow{2}{*}{Mistral-7B}       & Zero        & $0.002 \pm 0.0006$      & $0.082 \pm 0.0102$     & $0.061 \pm 0.0401$      & $0.211 \pm 0.0615$     \\
                              &                                & ICL         & $0.055 \pm 0.0183$      & $0.235 \pm 0.0408$     & $0.114 \pm 0.0290$      & $0.353 \pm 0.0470$     \\  \cmidrule(lr){3-7}
                              & GPT-3.5                         & ICL         & $0.041 \pm 0.0163$      & $0.252 \pm 0.0429$     & $0.121 \pm 0.0126$      & $0.365 \pm 0.0451$     \\ 
    \bottomrule
  \end{tabular}
  }
  \caption{Feedback generation performance on all settings with different data splits. Values are averages over all folds.}
  \label{tbl:split_results}
\end{table*}

\subsubsection{Experiment 1: Response-Level Split} \label{sec:response-ex}
In this experiment, we perform five-fold cross-validation across the unique erroneous student responses in the dataset. Specifically, we randomly sample 100 responses from the dataset to use for the train set and randomly sample 500 responses for the test set. We then use the training set both as examples to fine-tune the model and as a pool for in-context examples, if applicable. After fine-tuning, we then prompt the model to generate feedback on the test-set. 

We report the results of this experiment in the top block of Table \ref{tbl:split_results}, which shows the mean and standard deviation of the all metrics across all five splits. We see that the generated feedback is most similar to the reference feedback when the model is provided an in-context example from the same error class. This conclusion is expected, since feedback messages in this dataset are template-based, which means that ones corresponding to the same error are very similar and only vary in question-specific terminology. Therefore, the LLM easily learns to copy the structure of the in-context example and then simply replaces the terms specific to the in-context example with those of the current question.

We also see that, in general, fine-tuning the LLM makes the generated feedback more similar to the reference feedback compared to the base LLM. This result is not surprising since the fine-tuning process aligns the LLM to the problem-specific vocabulary after training on ground-truth feedback messages. However, we see that performance improvement is only minor after fine-tuning under the zero-shot setting. This observation highlights the importance of ICL as an efficient way to inform the LLM on the style of feedback compared to fine-tuning alone. 

Moreover, we see that GPT-3.5 outperforms Mistral-7B in all settings. This result is somewhat expected since GPT-3.5 is orders of magnitude larger than Mistral-7B. GPT-3.5 works relatively well even without fine-tuning as long as it has access to ICL examples via the prompt.


Furthermore, we see that ICL-SE significantly outperforms ICL. This observation suggests that when the right ICL example, i.e., one that contains feedback for the exact same student error as the student response at hand, is provided, LLMs can replicate the ITS's built-in feedback much more accurately compared to other ICL examples. Since feedback messages for the same error across different questions/responses share a common template, this result verifies the demonstration-following capability of LLMs in replicating output according to input instructions. 


\subsubsection{Experiment 2: Question-Level Split}
In this experiment, we perform five-fold cross validation across the unique questions in the dataset. First, we randomly perform an 80\%-20\% train-test split across the unique questions in the dataset. Second, we randomly sample 100 train and 500 test responses from the train and test splits respectively. We repeat this split-sampling process for each fold. The remainder of the experiment setup is identical to that in the previous experiment.

We report the results of this experiment in the middle block of Table \ref{tbl:split_results}. We observe similar trends in this experiment in comparison to the first experiment. The reason for this result is very likely the strong association between questions and feedback messages in the dataset, since these feedback messages only vary by a small number of tokens across questions. However, while the resulting trends are similar, the practical implication of this result is more compelling than that of the first experiment: Given a small set of examples, an LLM can generate reasonable feedback messages close to that of a hand-crafted template system for previously unseen questions. Therefore, ITS designers can potentially scale up their built-in feedback mechanism to a large number of unique questions with the help of LLMs.

\subsubsection{Experiment 3: Error-Level Split}\label{sec:erorr-ex}
In this experiment, we perform five-fold cross validation across the unique error classes in the dataset. First, we randomly select 2 of the 11 unique error classes and split our dataset into two groups. The split with the more distinct errors forms the train split and the group with less distinct errors other forms the test split. Second, we randomly sample 100 train and 500 test responses from the train and test splits respectively. We repeat this split-sampling process for each fold. The remainder of the experiment setup is then identical to the prior two experiments.

We report the results of this experiment in the bottom block of Table \ref{tbl:split_results}. We see a drastic decrease in performance across all settings, compared to the previous experiments, for both LLMs. 
This poor result is perhaps surprising since LLMs, especially GPT-3.5, have previously reported good mathematical reasoning ability, performing very well on math question answering tasks~\cite{frieder2023mathematical}. This observation suggests that while these LLMs may be able to answer questions, they do not seem to have be capable of understanding flawed reasoning exhibited by real students that leads to erroneous answers. This result draws into question whether LLMs are well-equipped to understand diverse reasoning strategies, which are required for feedback generation. It is likely that their performance in the feedback generation task mainly comes from demonstrations on the exact formatting of feedback, rather than truly understanding student errors. 

Table~\ref{tbl:abl_qual} shows the feedback messages generated for the same question-response pair in the test set, from different data splits. We see that when split by response, the generated feedback message is exactly the same, since the LLM has seen very similar examples during training. When split by questions, the LLM still generates good feedback although not using the exact terminology in the current question. When split by errors, the LLM generates a misleading feedback message suggesting it does not understand what feedback should look like for previously unseen student errors.


\begin{table}[t]
    \centering
\scalebox{.9}{
    \begin{tabular}{cp{70mm}}
        \toprule
         & Feedback \\
        \midrule
        \multirow{3}{*}{ITS} & You've got the right numbers, but think about what is constant and what changes as the number of holes the golfer plays increases. \\\cmidrule(lr){1-2}
        \multirow{3}{*}{R.\ Split} & You've got the right numbers, but think about what is the constant and what changes as the number of holes the golfer plays increases. \\\cmidrule(lr){1-2}
        \multirow{3}{*}{Q.\ Split} & Be careful with the starting amount and the way the number of balls changes as the amount of time increases. \\\cmidrule(lr){1-2}
        \multirow{2}{*}{E.\ Split} & Does the number of balls go up or go down as the number of holes the golfer plays increases? \\
        \bottomrule
    \end{tabular}
}
    \caption{Feedback messages generated from different data splits (response: R, question Q, and error: E)  for the same question-response pair in the test set, where we see feedback quality varying over each split.}
    \label{tbl:abl_qual}
\end{table}

\subsubsection{Experiment 4: Error-Information Ablation}
Following our observations in Experiment 3, we perform an additional experiment where we add the error label information included in the dataset. We take the best performing method from experiment 1, ICL-SE, and compare performance with or without error label information included in the prompt. We report the results in Table~\ref{tbl:ablate-err}. We see that the inclusion of the error label significantly improves the performance of feedback generation for GPT-3.5, but decreases performance for Mistral-7B. This result is surprising since intuitively, providing error labels to the LLM gives it specific information on the error in a student response, and should result in better feedback. This observation can likely be explained by the difference in scale and, perhaps by extension, intrinsic mathematical reasoning capabilities between these two models. Smaller models like Mistral-7B may get confused by the word overlap between error labels (e.g., ``forgot intercept'' and ``negated intercept''), while larger models like GPT-3.5 can at least parse the purpose of the labels and use them to guide feedback generation. Given these observations, we conclude that it is perhaps best to rely on LLMs for their text generation capabilities only, while finding other ways, such as template-based error detection and ICL demonstration, to inform them of student errors and feedback formatting.

\begin{table}
    \centering
\scalebox{0.9}{
    \begin{tabular}{clcc}
        \toprule
        Model & Variant & BLEU($\pm$STD) & ROUGE($\pm$STD) \\
        \midrule
        \multirow{2}{*}{Mistral-7B} & ICL-SE & $0.342 \pm 0.0496$ & $0.464 \pm 0.0470$\\
                                 & ICL-SE+EC & $0.254 \pm 0.0462$& $0.359 \pm 0.0487$\\
                                 \cmidrule(lr){2-4}
        \multirow{2}{*}{GPT-3.5} & ICL-SE & $0.509 \pm 0.0278$ & $0.700 \pm 0.0196$ \\
                                & ICL-SE+EC & $0.583 \pm 0.0169$ & $0.782 \pm 0.0131$\\ 
        \bottomrule
    \end{tabular}
}
    \caption{Ablation study with error label information in the response-wise split setting. +EC indicates error class included in the prompt.}
    \label{tbl:ablate-err}
\end{table}

\section{Conclusion and Future Work}

In this work, we conducted an examination on the capabilities of LLMs on generating feedback for open-ended math questions, using data from a real-world ITS. We experimented with a variety of common strategies for adapting LLMs to the task of feedback generation, using both open-source and proprietary LLMs. Our results indicate that LLMs show promise in replicating feedback messages which are similar to those shown during training, but struggle to generalize to previously unseen student errors. Our observations show that proprietary models such as GPT-3.5 outperform open-source models, such as Mistral-7B. We also find that the inclusion of explicit error label information can decrease performance of some models, which suggests that LLMs' ability to generate feedback comes fundamentally from instruction following rather than understanding student errors. 

There are many avenues for future work. First, comparing the performance between open-source and proprietary LLMs is not entirely fair due to their vast difference in scale. Therefore, experiments with larger open-source models, such as Llama-2 70b, remain to be performed, although setting them up can be challenging. Second, our findings suggest that error label text is not always an effective representation of student errors, which suggests that better representations, perhaps a latent one \cite{liu2023gptbased}, is worth further investigation. Third, one of the fundamental limitations of this study is the similarity between questions in the dataset since they are all under one single topic. Therefore, we may repeat our experiments on a set of more diverse open-ended math questions, both in terms of topics and question formats. This experiment may help us understand further the mathematical reasoning capabilities and limitations of LLMs. 

\section{Acknowledgements}
We thank Schmidt Futures and the NSF (under grants IIS-2118706 and IIS-2237676) for partially supporting this work. 

\newpage
\bibliographystyle{abbrv}
\bibliography{sigproc}  

\begin{thebibliography}{10}

\bibitem{socratic}
E.~Al-Hossami, R.~Bunescu, R.~Teehan, L.~Powell, K.~Mahajan, and M.~Dorodchi.
\newblock Socratic questioning of novice debuggers: A benchmark dataset and preliminary evaluations.
\newblock In {\em Proceedings of the 18th Workshop on Innovative Use of NLP for Building Educational Applications (BEA 2023)}, pages 709--726, Toronto, Canada, July 2023. Association for Computational Linguistics.

\bibitem{baral2023investigating}
S.~Baral, A.~F. Botelho, A.~Santhanam, A.~Gurung, J.~Erickson, and N.~T. Heffernan.
\newblock Investigating patterns of tone and sentiment in teacher written feedback messages.
\newblock In {\em International Conference on Artificial Intelligence in Education}, pages 341--346. Springer, 2023.

\bibitem{feedback-gen-neil}
A.~Botelho, S.~Baral, J.~A. Erickson, P.~Benachamardi, and N.~T. Heffernan.
\newblock Leveraging natural language processing to support automated assessment and feedback for student open responses in mathematics.
\newblock {\em Journal of Computer Assisted Learning}, 39(3):823--840, 2023.

\bibitem{brown2020language}
T.~B. Brown, B.~Mann, N.~Ryder, M.~Subbiah, J.~Kaplan, P.~Dhariwal, A.~Neelakantan, P.~Shyam, G.~Sastry, A.~Askell, S.~Agarwal, A.~Herbert-Voss, G.~Krueger, T.~Henighan, R.~Child, A.~Ramesh, D.~M. Ziegler, J.~Wu, C.~Winter, C.~Hesse, M.~Chen, E.~Sigler, M.~Litwin, S.~Gray, B.~Chess, J.~Clark, C.~Berner, S.~McCandlish, A.~Radford, I.~Sutskever, and D.~Amodei.
\newblock Language models are few-shot learners, 2020.

\bibitem{frieder2023mathematical}
S.~Frieder, L.~Pinchetti, A.~Chevalier, R.-R. Griffiths, T.~Salvatori, T.~Lukasiewicz, P.~C. Petersen, and J.~Berner.
\newblock Mathematical capabilities of chatgpt, 2023.

\bibitem{gurung2023common}
A.~Gurung, S.~Baral, M.~P. Lee, A.~C. Sales, A.~Haim, K.~P. Vanacore, A.~A. McReynolds, H.~Kreisberg, C.~Heffernan, and N.~T. Heffernan.
\newblock How common are common wrong answers? crowdsourcing remediation at scale.
\newblock In {\em Proceedings of the Tenth ACM Conference on Learning@ Scale}, pages 70--80, 2023.

\bibitem{haim2022toward}
A.~Haim, E.~Prihar, and N.~T. Heffernan.
\newblock Toward improving effectiveness of crowdsourced, on-demand assistance from educators in online learning platforms.
\newblock In {\em International Conference on Artificial Intelligence in Education}, pages 29--34. Springer, 2022.

\bibitem{hu2021lora}
E.~J. Hu, Y.~Shen, P.~Wallis, Z.~Allen-Zhu, Y.~Li, S.~Wang, L.~Wang, and W.~Chen.
\newblock Lora: Low-rank adaptation of large language models, 2021.

\bibitem{insta-reviewer}
Q.~Jia, M.~Young, Y.~Xiao, J.~Cui, C.~Liu, P.~Rashid, and E.~Gehringer.
\newblock Insta-reviewer: A data-driven approach for generating instant feedback on students' project reports.
\newblock {\em International Educational Data Mining Society}, 2022.

\bibitem{jiang2023mistral}
A.~Q. Jiang, A.~Sablayrolles, A.~Mensch, C.~Bamford, D.~S. Chaplot, D.~de~las Casas, F.~Bressand, G.~Lengyel, G.~Lample, L.~Saulnier, L.~R. Lavaud, M.-A. Lachaux, P.~Stock, T.~L. Scao, T.~Lavril, T.~Wang, T.~Lacroix, and W.~E. Sayed.
\newblock Mistral 7b, 2023.

\bibitem{feedback-effect-2}
E.~Kochmar, D.~D. Vu, R.~Belfer, V.~Gupta, I.~V. Serban, and J.~Pineau.
\newblock Automated personalized feedback improves learning gains in an intelligent tutoring system.
\newblock In {\em Artificial Intelligence in Education: 21st International Conference, AIED 2020, Ifrane, Morocco, July 6--10, 2020, Proceedings, Part II 21}, pages 140--146. Springer, 2020.

\bibitem{koutcheme2023training}
C.~Koutcheme.
\newblock Training language models for programming feedback using automated repair tools.
\newblock In {\em International Conference on Artificial Intelligence in Education}, pages 830--835. Springer, 2023.

\bibitem{lan2015mathematical}
A.~S. Lan, D.~Vats, A.~E. Waters, and R.~G. Baraniuk.
\newblock Mathematical language processing: Automatic grading and feedback for open response mathematical questions.
\newblock In {\em Proceedings of the second (2015) ACM conference on learning@ scale}, pages 167--176, 2015.

\bibitem{lin-2004-rouge}
C.-Y. Lin.
\newblock {ROUGE}: A package for automatic evaluation of summaries.
\newblock In {\em Text Summarization Branches Out}, pages 74--81, Barcelona, Spain, July 2004. Association for Computational Linguistics.

\bibitem{incontext2021gpt3}
J.~Liu, D.~Shen, Y.~Zhang, B.~Dolan, L.~Carin, and W.~Chen.
\newblock What makes good in-context examples for gpt-3.
\newblock abs/2101.06804, 2021.

\bibitem{liu2023gptbased}
N.~Liu, Z.~Wang, R.~G. Baraniuk, and A.~Lan.
\newblock Gpt-based open-ended knowledge tracing, 2023.

\bibitem{loshchilov2019decoupled}
I.~Loshchilov and F.~Hutter.
\newblock Decoupled weight decay regularization, 2019.

\bibitem{distractors}
H.~McNichols, W.~Feng, J.~Lee, A.~Scarlatos, D.~Smith, S.~Woodhead, and A.~Lan.
\newblock Automated distractor and feedback generation for math multiple-choice questions via in-context learning.
\newblock {\em NeurIPS’23 Workshop on Generative AI for Education}, 2023.

\bibitem{mcnichols2023exploring}
H.~McNichols, W.~Feng, J.~Lee, A.~Scarlatos, D.~Smith, S.~Woodhead, and A.~Lan.
\newblock Exploring automated distractor and feedback generation for math multiple-choice questions via in-context learning.
\newblock {\em arXiv preprint arXiv:2308.03234}, 2023.

\bibitem{feedback-gen-decimal}
H.~A. Nguyen, H.~Stec, X.~Hou, S.~Di, and B.~M. McLaren.
\newblock Evaluating chatgpt's decimal skills and feedback generation in a digital learning game.
\newblock In {\em Responsive and Sustainable Educational Futures}, pages 278--293, Cham, 2023. Springer Nature Switzerland.

\bibitem{chatgpt}
OpenAI.
\newblock Introducing chatgpt, 2022.

\bibitem{papineni2002bleu}
K.~Papineni, S.~Roukos, T.~Ward, and W.-J. Zhu.
\newblock Bleu: a method for automatic evaluation of machine translation.
\newblock In {\em Proceedings of the 40th annual meeting of the Association for Computational Linguistics}, pages 311--318, 2002.

\bibitem{prihar2023comparing}
E.~Prihar, M.~Lee, M.~Hopman, A.~T. Kalai, S.~Vempala, A.~Wang, G.~Wickline, A.~Murray, and N.~Heffernan.
\newblock Comparing different approaches to generating mathematics explanations using large language models.
\newblock In {\em International Conference on Artificial Intelligence in Education}, pages 290--295. Springer, 2023.

\bibitem{feedback-effect-1}
R.~Razzaq, K.~S. Ostrow, and N.~T. Heffernan.
\newblock Effect of immediate feedback on math achievement at the high school level.
\newblock In {\em International Conference on Artificial Intelligence in Education}, pages 263--267. Springer, 2020.

\bibitem{stamper2008hint}
J.~Stamper, T.~Barnes, L.~Lehmann, and M.~Croy.
\newblock The hint factory: Automatic generation of contextualized help for existing computer aided instruction.
\newblock In {\em Proceedings of the 9th International Conference on Intelligent Tutoring Systems Young Researchers Track}, pages 71--78, 2008.

\bibitem{feedback-gen-chatgpt}
J.~Steiss, T.~Tate, S.~Graham, J.~Cruz, M.~Hebert, J.~Wang, Y.~Moon, W.~Tseng, et~al.
\newblock Comparing the quality of human and chatgpt feedback on students’ writing.
\newblock 2023.

\bibitem{touvron2023llama}
H.~Touvron, L.~Martin, K.~Stone, P.~Albert, A.~Almahairi, Y.~Babaei, N.~Bashlykov, S.~Batra, P.~Bhargava, S.~Bhosale, et~al.
\newblock Llama 2: Open foundation and fine-tuned chat models.
\newblock {\em arXiv preprint arXiv:2307.09288}, 2023.

\end{thebibliography}

\end{document}